\documentclass[10pt,twocolumn,letterpaper]{article}

\usepackage{cvpr}              
\definecolor{cvprblue}{rgb}{0.21,0.49,0.74}
\usepackage[pagebackref,breaklinks,colorlinks,allcolors=cvprblue]{hyperref}

\def\paperID{3638} 
\def\confName{CVPR}
\def\confYear{2026}

\title{Physics-guided Image Dehazing Diffusion}

\author{
Shijun Zhou$^{1,2}$, Xing Xie$^{1,2}$, Baojie Fan$^{3}$, Jiandong Tian$^{1}$\\
$^{1}$State Key Laboratory of Robotics and Intelligent Systems\\
Shenyang Institute of Automation, Chinese Academy of Sciences\\
$^{2}$University of Chinese Academy of Sciences\\
$^{3}$Department of Automation, Nanjing University of Posts and Telecommunications\\
{\tt\small \{zhoushijun, xiexing\}@sia.cn, jobfbj@gmail.com, tianjd@sia.cn}
}

\begin{document}
\maketitle
\begin{abstract}
Due to the domain gap between real-world and synthetic hazy images, current data-driven dehazing algorithms trained on synthetic datasets perform well on synthetic data but struggle to generalize to real-world scenarios. To address this challenge, we propose \textbf{I}mage \textbf{D}ehazing \textbf{D}iffusion \textbf{M}odels (IDDM), a novel diffusion process that incorporates the atmospheric scattering model into noise diffusion. IDDM aims to use the gradual haze formation process to help the denoising Unet robustly learn the distribution of clear images from the conditional input hazy images. 
During the forward process, IDDM simultaneously introduces haze and noise into clear images, and then robustly separates them during the sampling process.
By training with physics-guided information, IDDM shows the ability of domain generalization, and effectively restores the real-world hazy images despite being trained on synthetic datasets. Extensive experiments demonstrate the effectiveness of our method through both quantitative and qualitative comparisons with state-of-the-art approaches.\end{abstract}    
\section{Introduction}
\label{sec:intro}
Due to the attenuation of light in the atmosphere, images captured in hazy conditions often suffer from blurriness. These degradations not only affect visual quality but also significantly impair the performance of computer vision systems in critical applications such as autonomous driving and surveillance. To address the obvious harm to machine vision systems, extensive research \cite{hazelines,dehazenet,nara2003,nara2000,NarasimhanNSK05,he_single_nodate} has been conducted in recent years on image dehazing.

      

\begin{figure}
    \centering
    \includegraphics[width=\linewidth]{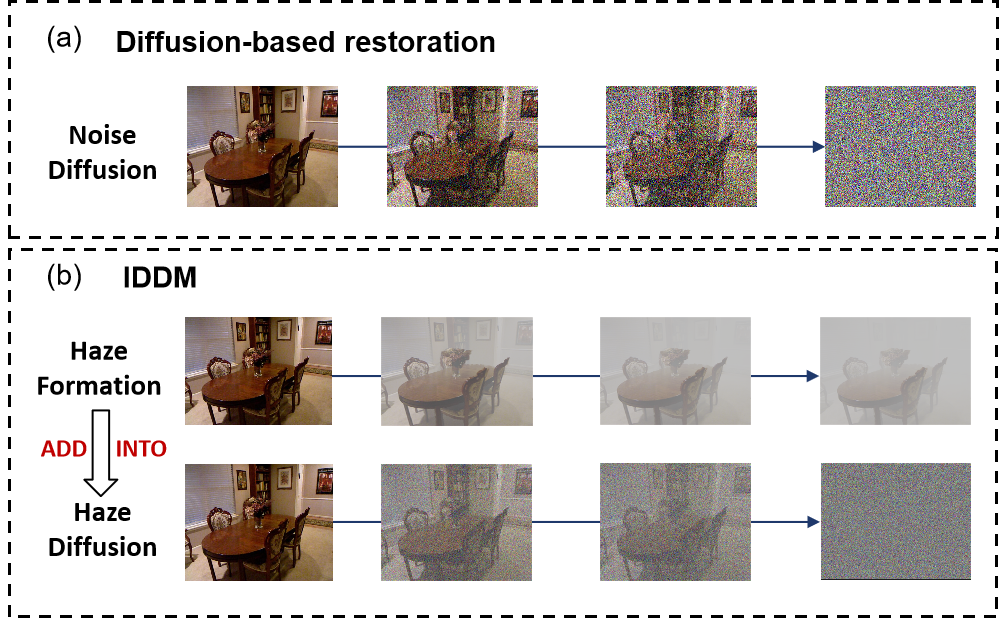}
    \caption{Motivation for physics-guided Dehazing diffusion. Haze formation and the diffusion forward process share a common characteristic: both progressively degrade image clarity. IDDM integrates the physical haze formation process (depth-indexed) into diffusion (time-indexed) by simultaneously adding atmospheric scattering and noise.}
    \label{fig:placeholder}
\end{figure}
  
Prior-based methods \cite{he_single_nodate,hazelines,fattal,tan,zhou,zhoulight} frequently lack robustness in scenarios where prior information is unavailable. In contrast, data-driven deep learning methods \cite{c2pnet, dehazenet,fang2025guided,fan2024driving,fan2025depth,FD-GAN} have proven to be more effective in handling complex environmental conditions. Current deep learning approaches for image dehazing \cite{Zhang_2024_CVPR,OKNET} are often trained on synthetic datasets \cite{reside, haze4k} with paired hazy and haze-free images, as acquiring real-world paired hazy data is challenging. However, models trained exclusively on synthetic data often exhibit poor generalization capabilities when applied to real-world images, especially under complex and variable weather conditions. This domain gap significantly limits the practical utility of these algorithms in real-world applications. Recent works have attempted to address this limitation through domain adaptation \cite{DAD}, physical priors \cite{psd}, or contrastive learning \cite{Zheng_2023_CVPR}, but the fundamental challenge remains.

The generalization challenge primarily stems from two issues: (1) the significant domain discrepancy between synthetic and real-world hazy images in terms of haze distribution, color characteristics, and scene complexity; and (2) the focus of previous models on learning the direct mapping from synthetic hazy images to clear ones, rather than learning the physical principles of haze formation, particularly how haze varies with airlight, scattering coefficients and depth. These limitations prevent models from developing a comprehensive understanding of the physical process underlying haze formation.




To address these challenges, models need to be gradually trained on the entire process of haze development—from clear conditions to dense haze—in a step-by-step manner. This approach contrasts with previous models, which treat the problem as a direct mapping between a hazy image and its clear counterpart without accounting for the intermediate stages of haze formation. By understanding the progressive nature of haze accumulation, models can develop more robust and physically consistent dehazing capabilities.
To implement this progressive learning approach, we turn to diffusion models, which have demonstrated remarkable success in various image generation and restoration tasks. We find that diffusion models offer an ideal foundation for this setting for several compelling reasons. The denoising diffusion models, which also contain inherent progressive learning strategies through their step-by-step noise addition and removal processes, excel in data distribution modeling, which can address the cross-domain challenge between synthetic and real images by learning the natural distribution of clear images. Moreover, there exists a remarkable parallel between the forward process of DDPM \cite{DDPM} and the physical process of haze formation in real-world scenarios. In a forward diffusion process, a clear image gradually becomes pure noise over time, while in the haze formation process, it gradually accumulates haze as depth increases. Both processes share a similar trajectory: the image progressively deteriorates, either into noisy or hazy. This natural correspondence provides a theoretical basis for integrating physical models into generative frameworks and makes diffusion models particularly suitable as a foundation for image dehazing.

However, while both haze formation and noise addition degrade image characteristics, they differ fundamentally in their physical mechanisms and spatial characteristics. Random noise affects each pixel independently, while haze formation involves physical interactions between atmospheric light and scene depth. Therefore, directly applying standard diffusion models to dehazing without incorporating physical principles would be insufficient for learning the underlying physical processes.

To overcome this limitation, we utilize their similarities to deduce a novel diffusion model (IDDM) specifically for image dehazing, which explicitly models the formation of haze through the diffusion process. 
We model haze as a timestep-dependent haze component $\mathbf{h}_t$ to integrate physical principles into the diffusion process. A unique bi-directional enhancement framework is created around it, enabling denoising and dehazing processes to mutually promote each other. $\mathbf{h}_t$ forms the foundation of our novel image dehazing diffusion model and serves as a conditional vector guiding the diffusion model's training, enabling accurate noise prediction across varying haze densities. 
Simultaneously, intermediate images with varying levels of noise and haze from the diffusion process are utilized to train the model to estimate haze under diverse conditions at each timestep. 
Our method demonstrates consistent improvements across multiple no-reference quality metrics on real-world datasets such as RTTS and BEDDE, validating its generalization ability to real-world conditions.


Notably, our formulation offers a key advantage over previous physics-guided methods. Unlike conventional physics-guided methods \cite{Varghese2023} that separately estimate atmospheric light and transmission—leading to ill-posed multi-variable problems—our approach directly predicts the unified haze component $\mathbf{h}_T$ and implicitly recovers the attenuated part, circumventing contradictory estimations.


The main contributions of this work can be summarized as follows:

(1) We introduce Image Dehazing Diffusion Models (IDDM), a novel framework that integrates the atmospheric scattering model into diffusion processes. This integration allows IDDM to simultaneously leverage the distribution learning capabilities of diffusion models and the physical constraints of atmospheric scattering, creating a physically-informed approach to dehazing.

(2) We propose a bidirectional training framework that mutually enhances diffusion model and haze estimator. The predicted haze component serves as conditioning to guide denosing process, while intermediate states generated by diffusion with varying noise-haze combinations provide step-wise supervision for training haze estimator.

(3) We evaluate the restoration performances on three real-world hazy image datasets, with state-of-the-art (SOTA) performances achieved.

\section{Related Works}
\label{sec:relatedworks}

\subsection{Single Image Dehazing}

Traditional image dehazing methods rely on hand-crafted priors such as dark channel \cite{he_single_nodate} and haze-lines \cite{hazelines}, which often struggle with complex scattering scenarios. The emergence of deep learning shifted this paradigm, with early CNN-based approaches \cite{dehazenet,ren2016,Zhang_2018_CVPR} learning to estimate transmission maps and atmospheric light, while later end-to-end methods \cite{Dong_2020_CVPR,Li_2017_ICCV,Liu_2019_ICCV,Qin_Wang_Bai_Xie_Jia_2020,Qu_2019_CVPR} directly mapped hazy to clear images. However, models trained purely on data-driven mappings without physical constraints often produce artifacts and exhibit poor generalization to real-world scenarios.

To address these limitations, recent physics-guided approaches \cite{Zheng_2023_CVPR,Wu_2023_CVPR,psd} integrate the atmospheric scattering model (ASM) to provide physical constraints during training. While these methods show improved performance, they typically require explicit estimation of multiple interdependent parameters (atmospheric light and transmission), leading to an ill-posed optimization problem. For real-world deployment, domain adaptation techniques \cite{DAD,D4,ridcp} have been proposed to bridge the synthetic-to-real gap, though challenges remain in handling diverse weather conditions and dense haze scenarios. These findings motivate our investigation of diffusion models as a unified framework that naturally captures both the physical haze formation process and the distribution of real-world images.

\subsection{Diffusion Models}

Diffusion probabilistic models \cite{pmlr-v37-sohl-dickstein15,DDPM} have emerged as powerful generative frameworks through their iterative denoising process, achieving remarkable success across diverse vision tasks including image restoration \cite{Xie_2024_CVPR,10021824,Garber_2024_CVPR}, super-resolution \cite{Gao_2023_CVPR}, deblurring \cite{Whang_2022_CVPR}, and content generation \cite{Rombach_2022_CVPR,Yang_2023_CVPR,Zhang_2023_CVPR}. Score-based formulations \cite{NEURIPS2019_3001ef25,meng2022sdedit,Chung_2022_CVPR} provide an alternative perspective, enabling flexible conditional generation through various guidance mechanisms. 

Despite their success in general image processing, applying diffusion models to physically-constrained restoration tasks like dehazing remains underexplored. Recent unified multimodal models \cite{blip3o} demonstrate the effectiveness of bi-directional conditioning between different model components, where intermediate representations mutually supervise each task. Inspired by these advances, we propose integrating physical haze formation into the diffusion framework, where the progressive nature of diffusion naturally aligns with atmospheric scattering processes. 

\begin{figure*}[t]
    \centering
    \includegraphics[width=\linewidth]{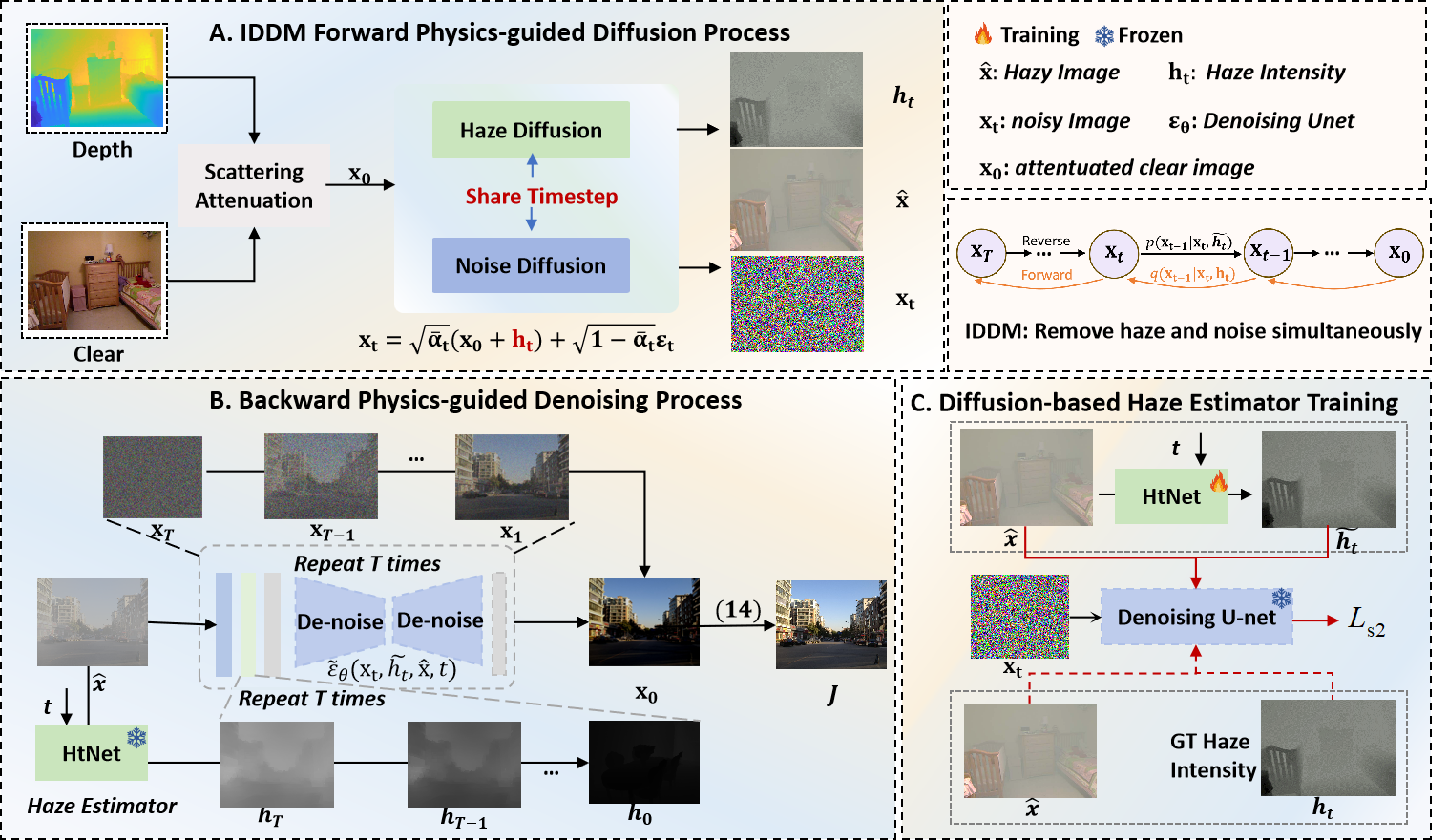}
    \caption{Overview of the proposed framework. We introduce a physics-guided diffusion process that explicitly models haze formation through the atmospheric scattering model. (A) During forward diffusion, both physical haze $\mathbf{h}_t$ and noise ${\boldsymbol{\epsilon}_t}$ are progressively added to clear images through shared timesteps, where $\mathbf{h}_t$ accumulates according to depth-dependent scattering attenuation. (B) The reverse process leverages a Haze Estimator (HtNet) to predict timestep-dependent haze intensity, which conditions the denoising network for physics-aware image restoration. (C) A bi-directional training strategy is employed: the frozen denoising U-Net supervises HtNet training via consistency regularization, while predicted haze components guide noise prediction across varying degradation levels.
}
    \label{fig:IDDM}
\end{figure*}
\section{Image Dehazing Diffusion Models}
\subsection{Why Diffusion for Dehazing}
Image dehazing aims to recover a clear image $J$ from its hazy observation $I$. According to the atmospheric scattering model (ASM) \cite{nara2000,nara2003}, this degradation process is described as:
\begin{equation}
\begin{aligned}
    I &= \underbrace{J e^{-\sigma z}}_{\text{Attenuation}} + \underbrace{\int_{0}^z A e^{-\sigma z} dz}_{\text{Haze Component}}\\
      &= \underbrace{J T_r}_{\text{Attenuation}} + \underbrace{A'(1-T_r)}_{\text{Haze Component}}.
\end{aligned}
    \label{asm}
\end{equation}

where $A = \frac{A'}{\sigma}, T_r = e^{-\sigma z}$. Here, $J$ denotes a clear image, $I$ denotes a hazy image, $A$ denotes the airlight, $z$ denotes depth, and $T_r$ denotes the transmission map. Here, a hazy image can be divided into two parts: attenuation and haze component. 

Base on Equation \ref{asm}, the recovery of the clear image $J$ necessitates two steps: calculating the Haze Component and inferring the transmission map $T_r$. 
A key observation is that the haze component accumulates progressively with depth -- as $z$ increases from 0 to infinity, the image gradually transitions from clear $J$ to completely hazy $A'$.

This progressive nature of haze formation shares a remarkable similarity with diffusion models, which also involve gradual transformation over time steps.  In diffusion models, a clear image progressively becomes noise through iterative addition. Given a data point sampled from a real data distribution $\mathbf{x}_0$, the forward diffusion process of DDPM at step $t$ can be explained as:
\begin{equation}
\label{diffusion forward process}
    \mathbf{x}_t =\sqrt{\bar{\alpha}_t} \mathbf{x}_0+\sqrt{1-\bar{\alpha}_t} \boldsymbol{\epsilon}_t,
\end{equation}
where $\boldsymbol{\epsilon}_t \sim \mathcal{N}(\mathbf{0}, \mathbf{I})$, $\bar{\alpha}_t=\prod_{i=1}^t \alpha_i$, and $\alpha_i$ denotes signal weight. In this equation, \(\mathbf{x}_t\) represents the data at time step \(t\) of diffusion, \(\alpha_t\) is the scaling factor that controls noise, and $\boldsymbol{\epsilon}_{t}$ is Gaussian noise sampled from \(\mathcal{N}(\mathbf{0}, \mathbf{I})\). Both processes share a fundamental similarity: progressive degradation through additive accumulation. While diffusion models add Gaussian noise $\boldsymbol{\epsilon}_t$ at each timestep, haze formation adds atmospheric scattering at each depth increment.

We ask: \textbf{\textit{Can we leverage this similarity to design a physics-guided diffusion model that naturally incorporates the haze formation process?}} Recent works \cite{domain1,zhang2023domain} have demonstrated the effectiveness of diffusion models for domain adaptation tasks, suggesting their inherent capability to learn transferable representations across different domains. This provides additional theoretical motivation for applying diffusion models to address the synthetic-to-real generalization challenge in image dehazing. Traditional dehazing methods either estimate $T_r$ and A separately or learn direct mappings, but they struggle with domain generalization. By integrating the physical process into the diffusion framework, we aim to learn not just the mapping but the underlying formation process, leading to better generalization.

\subsection{Modeling Haze Formation as a Diffusion Process}
\label{haze_formation}
Motivated by the analogy between progressive haze accumulation and diffusion noise addition, we reformulate haze formation as a time-indexed process that mirrors diffusion timesteps.

\noindent \textbf{From Depth-Dependent to Time-Dependent.}
In the atmospheric scattering model, the amount of haze observed at a pixel $\mathbf{h}(x,y)$ is determined by integrating atmospheric scattering along the light path from depth $0$ to the scene depth $Z(x,y)$:
\begin{equation}
\label{eq:xy}
 	\mathbf{h}(x,y) = A \int_0^{Z(x,y)} e^{-\sigma z} d z.
\end{equation}
\textbf{We reformulate this continuous, depth-dependent process as a discrete, time-dependent process.} For any pixel with scene depth $Z$, we can partition the total integration path $[0, Z]$ into $T$ sequential segments. The total haze $\mathbf{h}(Z)$ can then be expressed as the sum of the haze contributed by each segment $t$, where each segment has a depth increment of $\frac{Z}{T}$:
\begin{equation}
\label{eq:hT}
 	\mathbf{h}(Z)= A \sum_{t=1}^T \int_{\frac{t-1}{T} \cdot Z}^{\frac{t}{T} \cdot Z} e^{-\sigma z} d z.
\end{equation}
Since the sum of integrals equals the integral of the sum, this discrete formulation in Equation \eqref{eq:hT} is \textbf{mathematically equivalent} to the continuous formulation in Equation \eqref{eq:xy}.

\noindent \textbf{Per-Timestep Haze Integration.}
This $T$-step reformulation allows us to define the progressive state of haze accumulation. We define $\mathbf{h}_t$ as the total haze accumulated up to the $t$-th timestep, corresponding to the integration from depth $0$ to $\frac{t}{T} \cdot Z$:
\begin{equation}
    \mathbf{h}_t = \int_0^{\frac{t}{T} \cdot Z} A e^{-\sigma z} d z.
    \label{eq:ht}
\end{equation}
The \textbf{incremental haze} added between consecutive timesteps $t-1$ and $t$ (i.e., from depth $\frac{t-1}{T} \cdot Z$ to $\frac{t}{T} \cdot Z$) is:
\begin{equation}
\Delta \mathbf{h}_t = \mathbf{h}_t - \mathbf{h}_{t-1} 
= A \int_{\frac{t-1}{T} \cdot Z}^{\frac{t}{T} \cdot Z} e^{-\sigma z} d z.
\label{eq:deltah}
\end{equation} 
Consequently, the final haze $\mathbf{h}_T$ at depth $Z$ is the cumulative sum of all incremental haze components:
\begin{equation}
 	\mathbf{h}_T = \sum_{t=1}^T \Delta \mathbf{h}_t.
 	\label{accu}
\end{equation}
\noindent \textbf{The parallel with diffusion.}
The above formulation reveals that haze accumulation is inherently additive, similar to how noise accumulates in diffusion models. In the end, we transfer haze from a depth-dependent value into a time-dependent value of diffusion. Therefore, \textbf{both processes} progressively degrade image quality through the increase of \textbf{time step}; both involve accumulation over time, making them naturally compatible for unified modeling in a single framework.


\subsection{The IDDM Forward Process}

Based on Equation \ref{asm}, we reformulate the diffusion forward process to adhere to physical principles: we set the base state $\mathbf{x}_0$ as the attenuated clear image $JT_r$, and restore attenuation $T_r$ later based on $\mathbf{h}_t$.
The diffusion process then progressively adds the haze component $\mathbf{h}_T$ to $\mathbf{x}_0$, directly modeling the physical accumulation process $I = \mathbf{x}_0 + \mathbf{h}_T$. 

Our key design simultaneously conducts \textbf{haze diffusion} and \textbf{noise diffusion} through shared timesteps (Fig. \ref{fig:IDDM}A). At each step $t$, haze component $\mathbf{h}_t$ accumulates according to the atmospheric scattering model (Eq. \ref{eq:ht}), while Gaussian noise $\boldsymbol{\epsilon}_t$ is added following the standard diffusion schedule. This synchronization enables the model to learn the joint distribution of haze and noise at varying degradation levels.

However, while both diffusion and haze accumulation exhibit progressive accumulation over time, they follow different mathematical formulations. Standard diffusion applies a linear combination with coefficient $\sqrt{\bar{\alpha}_t}$, whereas haze accumulation accumulates $\Delta \mathbf{h}_t$ without it. Therefore, integrating haze into the diffusion framework requires careful mathematical formulation to ensure compatibility. To maintain mathematical consistency, we derive the appropriate scaling for incremental haze addition (detailed derivation in Appendix), and propose a modified forward process that incorporates haze in a diffusion-compatible manner:
\begin{equation}
\label{fusing}
    \begin{aligned}
    \mathbf{x}_t  =&\sqrt{\alpha_t} \mathbf{x}_{t-1}+\sqrt{1-\alpha_t} \boldsymbol{\epsilon}_{t-1} +\sqrt{\bar{\alpha}_t} \Delta \mathbf{h}_t \\
     =& \sqrt{\alpha_t} (\sqrt{\alpha_{t-1}} \mathbf{x}_{t-2} + \sqrt{1-\alpha_{t-1}} \boldsymbol{\epsilon}_{t-2} + \sqrt{\bar{\alpha}_{t-1}} \Delta \mathbf{h}_{t-1})+ \\
    & \sqrt{1-\alpha_t} \boldsymbol{\epsilon}_{t-1} + \sqrt{\bar{\alpha}_t} \Delta \mathbf{h}_t \\
     =&\ldots \\
     =&\sqrt{\bar{\alpha}_t} (\mathbf{x}_0+ \mathbf{h}_t) +\sqrt{1-\bar{\alpha}_t} {\boldsymbol{\epsilon}_t}.
    \end{aligned}
\end{equation}
Consequently, the distribution for the diffusion process we propose is described by:
\begin{equation}
\small
    q(\mathbf{x}_t | \mathbf{x}_{t-1}, \mathbf{h}_t) = N(\mathbf{x}_t; \sqrt{\alpha_t}\mathbf{x}_{t-1} + \sqrt{\bar{\alpha}_t} \Delta \mathbf{h}_t, \sqrt{1-\alpha_t} \boldsymbol{\mathbf{I}}),
\end{equation}
\begin{equation}
    q(\mathbf{x}_t | \mathbf{x}_{0}, \mathbf{h}_t) = N(\mathbf{x}_t; \sqrt{\bar{\alpha}_t} (\mathbf{x}_{0}+ \mathbf{h}_t), \sqrt{1-\bar{\alpha}_t} \boldsymbol{\mathbf{I}}).
\end{equation}
In the end, haze of varying densities is combined with noise and fed into the model over $T$ steps. During the forward training process, $\mathbf{h}_t$ serves as a conditional vector and a haze component. This enables the model to learn how clear images and haze gradually combine, allowing it to remove haze during inference physically. Theoretical analyses, including forward process convergence, variational lower bound derivation, and the ELBO formulation, are provided in the \textbf{appendix} for completeness.

\subsection{The IDDM Sampling Process}
\label{sec:samp}

The determinisic implicit sampling process of IDDM can be derived by referring DDIM \cite{DDIM}, shown in Fig. \ref{fig:IDDM}. 
\begin{equation}
\label{hazeddim}
\begin{aligned}
\mathbf{x_{t-1}}=&\sqrt{\bar{\alpha}_{t-1}}\left(\mathbf{x}_0 + \mathbf{h}_t\right)+\sqrt{1-\bar{\alpha}_{t-1}-\sigma_t^2} \cdot \boldsymbol \epsilon_\theta^{(t)}+\sigma_t \boldsymbol \epsilon_\theta^{(t)} \\
=&\sqrt{\bar{\alpha}_{t-1}}\left(\frac{\mathbf{x}_t-\sqrt{1-\bar{\alpha}_t} \boldsymbol  \epsilon_\theta^{(t)}}{\sqrt{\bar{\alpha}_t}}\right)\\
&+\sqrt{1-\bar{\alpha}_{t-1}-\sigma_t^2} \cdot \boldsymbol{\epsilon}_\theta^{(t)}+\sigma_t \boldsymbol \epsilon_\theta^{(t)} - \sqrt{\bar{\alpha}_{t-1}}\Delta \mathbf{h}_t,
\end{aligned}
\end{equation}
where $\boldsymbol \epsilon_\theta^{(t)} = \boldsymbol \epsilon_\theta \left(\mathbf{x}_t, \mathbf{h}_t, \mathbf{\hat x}, t\right)$. Then based on Eq. \eqref{hazeddim}, computational cost can be reduced by utilizing only a subset of timestep indices, which is the same as DDIM \cite{DDIM}. Instead of considering all timesteps from $1$ to $T$, a subsequence $(\tau_1, \tau_2, \ldots, \tau_s)$ is selected. This subsequence is determined by evenly spacing out indices from the complete sequence $(1, 2, \ldots, T)$ using the formula $T_i = (i - 1) \cdot \frac{T}{S} + 1$, where $i$ ranges from $1$ to $\tau_s$ and $s$ is a parameter controlling the sampling rate. Therefore, for step 
\(\tau_1 < \tau_2\), the sampling process is given by
\begin{equation}
\begin{aligned}
\mathbf{x}_{\tau_1}=&\sqrt{\bar{\alpha}_{\tau_1}}\left( \mathbf{x}_0 - (\mathbf{h}_{\tau_2} - \mathbf{h}_{\tau_1})\right)\\&+\sqrt{1-\bar{\alpha}_{\tau_1}-\sigma_{\tau_2}^2} \cdot \boldsymbol \epsilon_\theta^{(\tau_2)}
+\sigma_{\tau_2} \boldsymbol \epsilon_\theta^{(\tau_2)}, 
\end{aligned}
\end{equation}
where
\begin{equation}
\mathbf{x}_0 = \frac{\mathbf{x}_{\tau_1}-\sqrt{1-\bar{\alpha}_{\tau_1}} \boldsymbol \epsilon_\theta^{({\tau_1})}}{\sqrt{\bar{\alpha}_{\tau_1}}},
\end{equation}
and a deterministic implicit sampling behavior sets \(\sigma_t^2 = 0\).

The predicted $\mathbf{x}_0$ from the diffusion process corresponds to the attenuated clear image ($J T_r$), while the haze component $\mathbf{h}_T$ is utilized to infer the transmission map $T_r$. The final clear image $J$ is recovered based on the Atmospheric Scattering Model (ASM) as $J = \mathbf{x}_0 / T_r$. Crucially, to simplify the restoration and avoid the need for explicit atmospheric light ($A'$) estimation, we leverage the property that the haze component $\mathbf{h}_T$ is proportional to $A'$. We therefore implicitly derive the transmission map $T_r$ by normalizing the estimated $\mathbf{h}_T$ at the final stage. The final restoration is computed as:
\begin{equation}
J = \frac{\mathbf{x}_0}{1 - \mathbf{\tilde{h}}_T},
\label{eq:final_restoration}
\end{equation}
where $\mathbf{\tilde{h}}_T$ is the stabilized and normalized haze component, representing $\mathbf{h}_T / A'$. This $\mathbf{\tilde{h}}_T$ is obtained by first applying a Gaussian filter to smooth the predicted $\mathbf{h}_T$, and then normalizing its pixel values to $[0, 1]$. This procedure effectively decouples the haze component from $A'$ and leads to superior visual quality. Prior multi-component estimation approaches \cite{Varghese2023} often require dedicated networks or separate modules to estimate $T_r$, $A$, and their combination individually. Conversely, we simplifies the restoration task by focusing solely on the direct synthesis of the haze component $\mathbf{h}_t$, contributing to a more efficient and robust restoration process.
\section{Method}

\begin{figure*}[t]
    \centering
    \includegraphics[width=\linewidth]{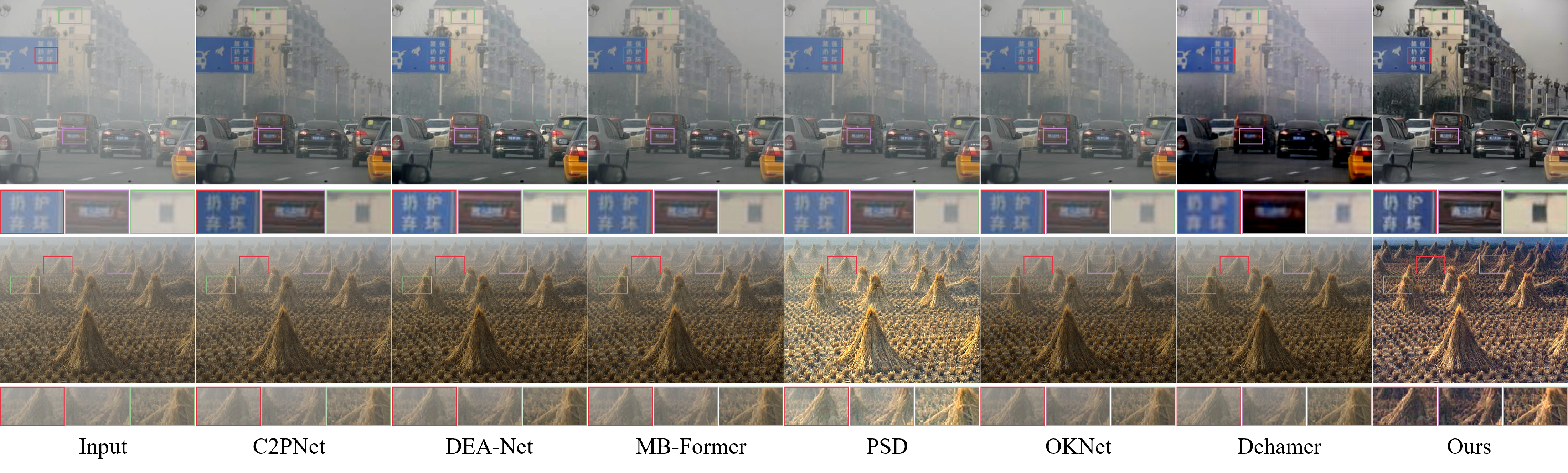}
    \caption{Visual comparisons on dense hazy images from RTTS dataset \cite{reside} and Fattal's dataset \cite{fattal}. All models are trained on OTS dataset. Please zoom in for more details.}
    \label{fig:OTS}
\end{figure*}

\begin{figure*}[t]
    \centering
    \includegraphics[width=\linewidth]{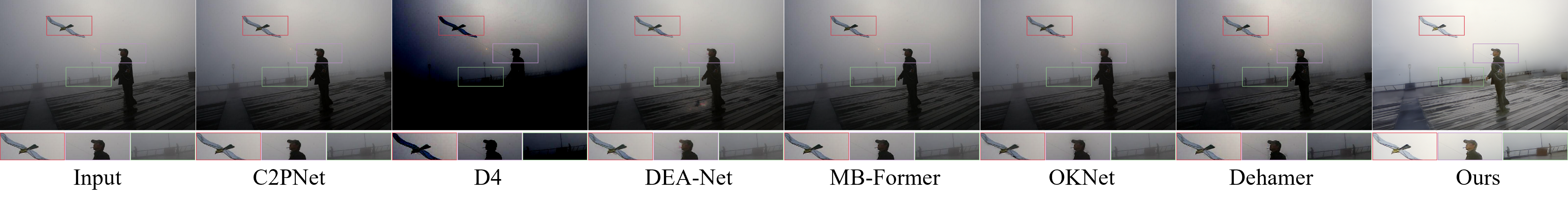}
    \caption{Visual comparisons on a low-light hazy image from RTTS dataset \cite{reside}. All models are trained on ITS dataset. }
    \label{fig:ITS}
\end{figure*}


Our training framework implements a two-stage sequential strategy that leverages the physical principles of IDDM's forward process. This design enables the network to jointly perform physics-guided dehazing and distribution-aware denoising during inference.



\subsection{Stage I: Physics-guided Denoising Training}
The first stage trains a denoising U-Net $\boldsymbol{\tilde\epsilon}_{\theta}$ to predict noise $\boldsymbol{\epsilon}_t$ from the diffused state $\mathbf{x}_t$, conditioned on both the ground-truth haze component $\mathbf{h}_t$ and the target hazy image $\hat{\mathbf{x}}$ (Fig. \ref{fig:IDDM}a). The training objective follows the standard diffusion loss:
\begin{equation}
\mathcal{L}_{s_1}= \mathbb{E}_{t \sim[1, T], \mathbf{x}_0 \sim q\left(\mathbf{x}_0\right)}\left\|\boldsymbol{\epsilon}_t-\boldsymbol{\tilde\epsilon_{\theta}}\left(\mathbf{x}_t, \mathbf{h}_t,\hat{\mathbf{x}}, t\right)\right\|^2,
\end{equation}
where $\mathbf{x}_t$ is computed via the IDDM forward process (Eq. \eqref{fusing}). By conditioning on $\mathbf{h}_t$, the model learns to denoise under varying physical haze densities, establishing the foundation for physics-aware restoration.


\subsection{Stage II: Diffusion-based Haze Estimator Training}
For real-world degraded images where ground-truth $\mathbf{h}_t$ is unavailable, we train a lightweight haze estimator (HtNet) to predict $\tilde{\mathbf{h}}_t$ from the input $\tilde{\mathbf{x}}$ at any timestep $t$. HtNet shares the U-Net architecture of Stage I but with reduced capacity (1/3 input channels). 

Critically, HtNet training is supervised by both ground-truth haze and the frozen Stage I network $\boldsymbol{\tilde\epsilon}_{\theta}$, creating a bi-directional enhancement mechanism (Fig. \ref{fig:IDDM}c). The dual-constraint objective ensures: (1) predicted haze matches ground-truth, and (2) noise prediction using estimated haze aligns with that using ground-truth haze:
\begin{equation}
\label{loss_s2}
    \mathcal{L}_{s_2}=  \left\|\mathbf{h}_t - \mathbf{\tilde{h}}_t\right\|_1 + \left\|\boldsymbol{\tilde\epsilon_{\theta}} \left(\mathbf{x}_t, \mathbf{\tilde{h}}_t,\hat{\mathbf{x}}, t\right) - \boldsymbol{\tilde\epsilon_{\theta}} \left(\mathbf{x}_t, \mathbf{h}_t,\hat{\mathbf{x}}, t\right)\right\|_1.
\end{equation}
This formulation enables HtNet to learn physically meaningful haze representations across all timesteps, while the frozen diffusion model acts as a consistency regularizer. By learning timestep-dependent haze concentrations, the model progressively removes haze over $T$ iterations during inference, naturally aligning with the physical accumulation process.

\begin{table*}[ht]
    \centering
    \small
    \renewcommand{\arraystretch}{0.9}
  \begin{tabular}{c|cccc|cccc}
  \hline
  Training Dataset & \multicolumn{4}{c|}{OTS} & \multicolumn{4}{c}{ITS} \\ \hline
  Criterion & Brisque$\downarrow$ & NIQE$\downarrow$ & PIQE$\downarrow$ & NIMA$\uparrow$ &
  Brisque$\downarrow$ & NIQE$\downarrow$ & PIQE$\downarrow$ & NIMA$\uparrow$ \\ \hline
  Dehamer \cite{dehamer} & 33.8582 & 4.9134 & 46.5483 & 4.6256 & {29.3234} & {4.5819} & {45.7394} & 4.5998 \\
  DEA-Net \cite{chen2023dea} & 31.1197 & 4.9206 & 48.1424 & 4.6499 & 29.7903 & 4.8192 & 45.813 & 4.6635 \\
  MB-TaylorFormer \cite{mbtaylor} & 32.6807 & 4.8664 & 46.8968 & 4.6184 & 32.8450 & 4.9090 & 47.7943 & 4.6302
   \\
  C2PNet \cite{c2pnet}& 34.2951 & 5.0336 & 48.6310 & 4.6399 & 33.7423 & 5.0546 & 48.5107 & {4.6739} \\
  PSD \cite{psd} & {25.2394} & {3.8742} & {30.5665} & 4.3459 & N/A & N/A & N/A & N/A \\
  D4 \cite{D4} & N/A & N/A & N/A & N/A & 33.206 & 5.8548 & 57.7313 & 3.7239 \\
  OKNet \cite{OKNET} & 35.1734 & 5.0858 & 49.1408 & {4.6548} & 33.1591 & 4.9664 & 49.1645 & 4.6558 \\
  \textbf{IDDM} & \textbf{18.3210} & \textbf{3.4937} & \textbf{27.0719} & \textbf{4.8401} & \textbf{27.8151}
  & \textbf{4.2198} & \textbf{38.9139} & \textbf{4.9153} \\ \hline
  \end{tabular}%
  \caption{The quantitative results of our method and the state-of-the-arts on real-world haze dataset RTTS.
  The best results are shown in bold.}
  \label{table1}
  \end{table*}

\begin{table}[t]
    \centering
    \small
    \renewcommand{\arraystretch}{0.8}
    \setlength{\tabcolsep}{4pt}
    \begin{tabular}{@{}l|ccc|ccc@{}}
    \toprule
    Method & \multicolumn{3}{c|}{OTS} & \multicolumn{3}{c}{ITS} \\
    \cmidrule(l){2-4} \cmidrule(l){5-7}
    & {VI$\uparrow$} & {RI$\uparrow$} & {VSI$\uparrow$} & {VI$\uparrow$} & {RI$\uparrow$} & {VSI$\uparrow$} \\
    \midrule
    Dehamer & 0.795 & 0.950 & 0.909 & 0.805 & 0.942 & {0.909} \\
    DEA-Net & {0.882} & 0.973 & 0.949 & 0.840 & 0.934 & 0.899 \\
    MB-TaylorFormer & 0.793 & 0.933 & 0.898 & 0.793 & 0.933 & 0.898 \\
    C2PNet & 0.880 & 0.970 & {0.948} & 0.795 & 0.913 & 0.874 \\
    D4/PSD & 0.860 & 0.966 & 0.925 & 0.814 & {0.951} & 0.923 \\
    OKNet & 0.875 & \textbf{0.973} & 0.947 & {0.846} & 0.935 & 0.908 \\
    \midrule
    IDDM (Ours) & \textbf{0.903} & {0.973} & \textbf{0.956} & \textbf{0.879} & \textbf{0.959} & \textbf{0.936} \\
    \bottomrule
    \end{tabular}
    \caption{Quantitative comparison on BEDDE dataset. The best results are shown in bold.}
    \label{tab:bedde}
\end{table}

\begin{table}[t]
  \centering
  \small
  \renewcommand{\arraystretch}{0.8}
  \setlength{\tabcolsep}{2pt}
  \begin{tabular}{@{}l|ccc|cccc@{}}
  \toprule
  Method & \multicolumn{3}{c|}{BEDDE} & \multicolumn{4}{c}{RTTS} \\
  \cmidrule(l){2-4} \cmidrule(l){5-8}
  & VI  & RI  & VSI  & Brisque  & NIQE & PIQE  & NIMA  \\
  \midrule
  RDDM & 0.778 & 0.957 & 0.908 & 22.391 & 6.876 & 43.704 & 4.270 \\
  Diff-Plugin & 0.785 & 0.956 & 0.908 & 25.651 & 4.613 & 28.339 & 4.915 \\
  \midrule
  RIDCP & 0.864 & 0.967 & 0.938 & 18.782 & 4.157 & 28.254 & 4.427 \\
  IPC-Dehaze & 0.877 & 0.970 & 0.943 & 19.447 & 4.084 & 31.644 & 4.865 \\
  \midrule
  Ours & {\textbf{0.878}} & {\textbf{0.979}} & {\textbf{0.946}} & {\textbf{17.734}} & {\textbf{3.7637}} & {\textbf{26.771}} & {\textbf{4.929}} \\
  \bottomrule
  \end{tabular}
  \caption{Comparison with two diffusion-based methods and two real-world dehazing methods on BEDDE and RTTS datasets. All methods are trained on the RIDCP dataset.}
  \label{tab:ridcp}
\end{table}

\section{Experiments}
\subsection{Experimental Settings}
\textbf{Training Datasets.} During training, we use ITS (13,990 indoor synthetic hazy image pairs) and OTS (72,135 outdoor synthetic hazy image pairs) datasets from RESIDE \cite{reside}. ITS depth maps come from NYU Depth Dataset V2 \cite{Silberman:ECCV12}, while OTS provides its own depth maps. We implement an online data generation strategy during training. For each clear image and corresponding depth map, we synthesize both the hazy image $\hat{\mathbf{x}}$ and the corresponding $\mathbf{h}_t$ at time step $t$ using the Atmospheric Scattering Model (ASM). 
Parameters such as airlight $A \in [0.7, 1.0]$ and scattering coefficient $\sigma \in [0.4, 1.5]$ are randomly sampled to guarantee data diversity and robustness across various haze conditions.

\noindent\textbf{Testing Datasets.} We qualitatively and quantitatively evaluate our method against state-of-the-art techniques on RTTS \cite{reside} (4,322 real-world hazy images) and BEDDE \cite{Bedde} (209 paired real-world images) datasets, with additional visual comparisons using Fattal's dataset \cite{fattal}.

\noindent \textbf{Implementation Details.}
During the training stages, experiments are conducted on 4 NVIDIA GeForce RTX 4090 GPUs. We optimize our networks with Adam optimizer ($\beta_1 = 0.9, \beta_2 = 0.99 $). The learning rate is set to 0.00001 and the batch size is set to 16. For data augmentation, we apply random horizontal and vertical flips, crop images to 256×256 pixels, and randomly introduce JPEG compression artifacts, gamma adjustment, and colorful haze based on \cite{Wu_2023_CVPR}. At stages 1, the diffusion denoising Unet is trained for 500,000 iterations, with the parameter $T$ is set as 1000. At stage 2, HtNet is trained for 60,000 iterations. Finally, all the evaluations for all methods are conducted on a NVIDIA GeForce RTX 3090 GPU.





\begin{figure}[t]
    \centering
    \begin{subfigure}[b]{0.32\linewidth}
        \includegraphics[width=\linewidth, height = 4pc]{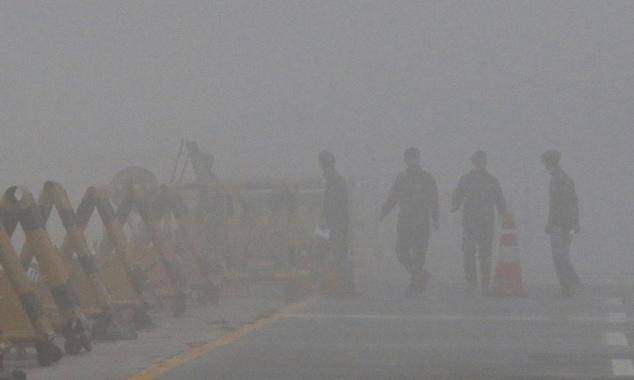}        
        \caption{Input}
    \end{subfigure}
    \hfill
    \begin{subfigure}[b]{0.32\linewidth}
        \includegraphics[width=\linewidth, height = 4pc]{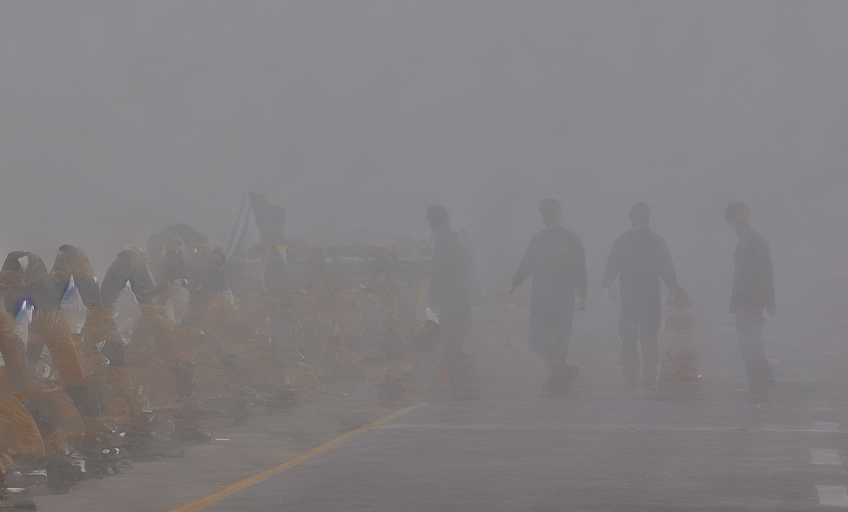}     
        \caption{Diff-Plugin}
    \end{subfigure}
    \hfill
    \begin{subfigure}[b]{0.32\linewidth}
        \includegraphics[width=\linewidth, height = 4pc]{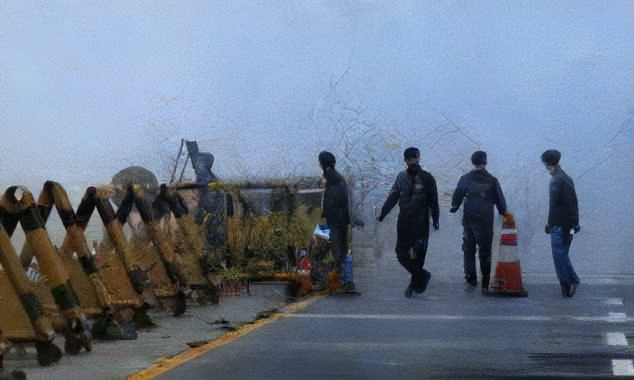} 
        \caption{IPC-Dehaze}
    \end{subfigure}
    \hfill
    \begin{subfigure}[b]{0.32\linewidth}
        \includegraphics[width=\linewidth, height = 4pc]{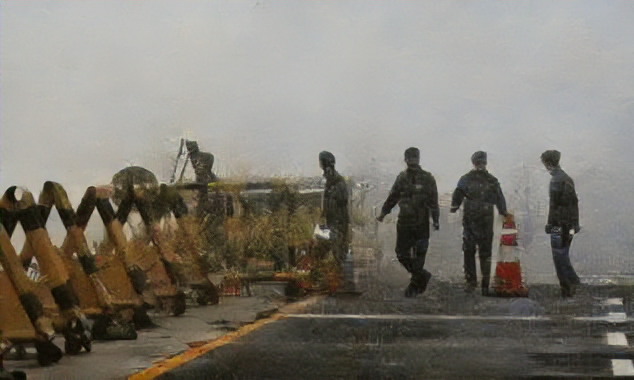}  
        \caption{RIDCP}
    \end{subfigure}
    \hfill
    \begin{subfigure}[b]{0.32\linewidth}
        \includegraphics[width=\linewidth, height = 4pc]{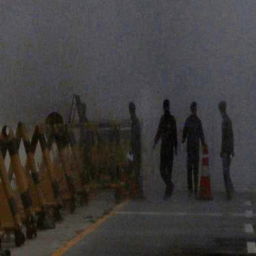}    
        \caption{RDDM}
    \end{subfigure}
    \hfill
    \begin{subfigure}[b]{0.32\linewidth}
        \includegraphics[width=\linewidth, height = 4pc]{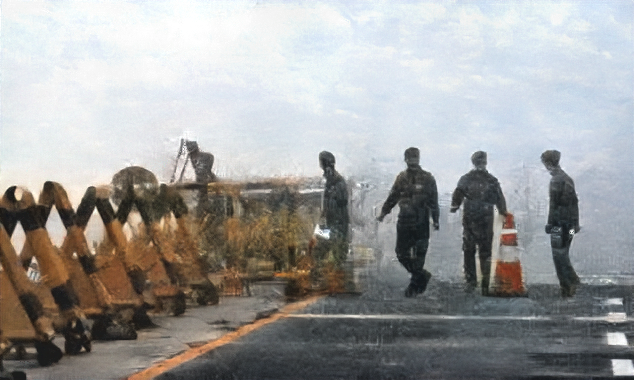}        
        \caption{Ours}
    \end{subfigure}
    \caption{Visual comparison of methods from Table. \ref{tab:ridcp} on a very dense hazy image from RTTS dataset \cite{reside}.}
    \label{fig:ridcp}
\end{figure}

\subsection{Results and Comparisons}

\noindent\textbf{Quantitative Comparison.} 
To evaluate the robustness of domain generalization, we train our models on both indoor and outdoor synthetic hazy datasets from RESIDE \cite{reside} and test them on two real-world datasets: RTTS \cite{reside} and BEDDE \cite{Bedde}.
We conduct comprehensive comparisons with state-of-the-art approaches: end-to-end deep learning methods, (Dehamer \cite{dehamer}, DEA-Net \cite{chen2023dea}, MB-TaylorFormer \cite{mbtaylor}, C2PNet \cite{c2pnet}, PSD \cite{psd}, D4 \cite{D4}, OKNET \cite{OKNET}), two real-world image dehazing methods which employ different training datasets beyond the standard ITS or OTS (RIDCP \cite{Wu_2023_CVPR}, IPC-Dehaze \cite{IPCDehaze}), and two diffusion-based approaches (Diff-Plugin \cite{diffplugin}, RDDM \cite{RDDM}). 
Notably, driven by the requirements of their methods, PSD \cite{psd} is trained exclusively on outdoor hazy images, while D4 \cite{D4} is trained solely on indoor hazy images.
As shown in Tables \ref{table1}, \ref{tab:bedde}, and \ref{tab:ridcp}, the experimental results demonstrate that our IDDM consistently outperforms all competing approaches, achieving the best performance across seven evaluation metrics.

\noindent \textbf{Qualitative Comparison.} As shown in Fig. \ref{fig:OTS} and Fig. \ref{fig:ITS}, we compare the visual results of our model with previous state-of-the-art methods on the RTTS dataset \cite{reside} and Fattal's data \cite{fattal}. In Fig. \ref{fig:OTS}, all models are trained on outdoor synthetic dataset, while in Fig. \ref{fig:ITS}, all models are trained on the indoor synthetic dataset. As shown in Fig. \ref{fig:ITS}, even when the training and test sets span two domains (indoor to outdoor, synthetic to real), IDDM still shows bright and clear dehazing performance. In contrast, the other methods tend to produce dark tones in the dehazed images, likely due to being trained exclusively on indoor hazy images.


\begin{figure}[t]
    \centering
    \begin{subfigure}[b]{0.32\linewidth}
        \includegraphics[width=\linewidth]{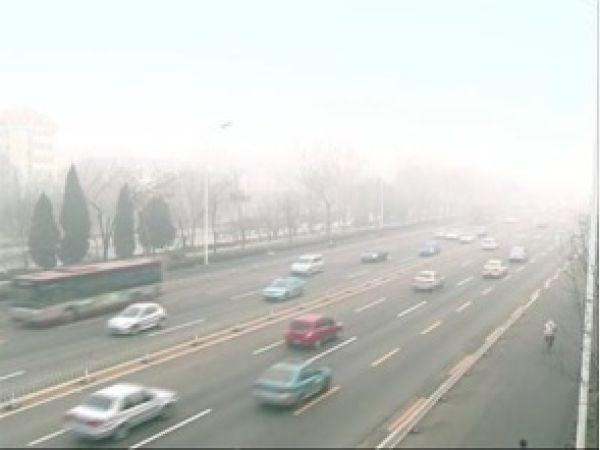}        
        \caption{Input}
    \end{subfigure}
    \hfill
    \begin{subfigure}[b]{0.32\linewidth}
        \includegraphics[width=\linewidth]{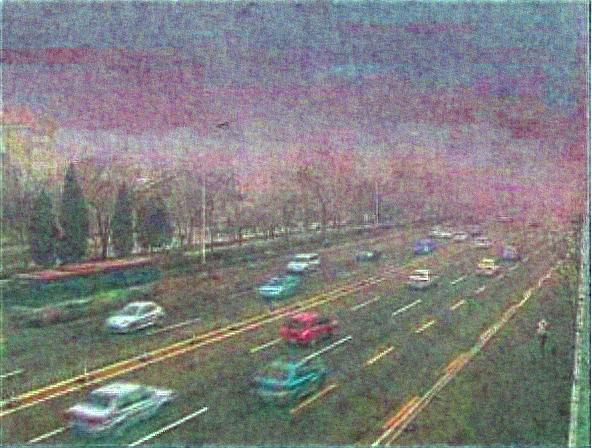}     
        \caption{w/o IDDM}
    \end{subfigure}
    \hfill
    \begin{subfigure}[b]{0.32\linewidth}
        \includegraphics[width=\linewidth]{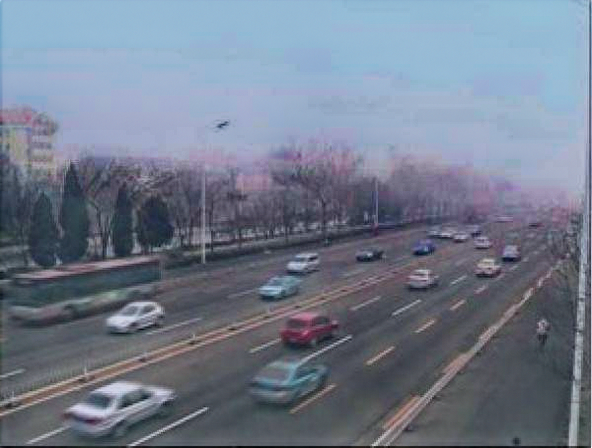} 
        \caption{w/o $\mathbf{h}_t$}
    \end{subfigure}
    \hfill
    \begin{subfigure}[b]{0.32\linewidth}
        \includegraphics[width=\linewidth]{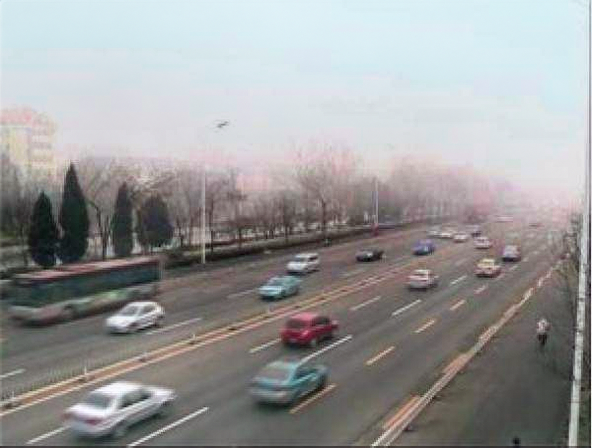}  
        \caption{w/o Eq. \ref{eq:final_restoration}}
    \end{subfigure}
    \hfill
    \begin{subfigure}[b]{0.32\linewidth}
        \includegraphics[width=\linewidth]{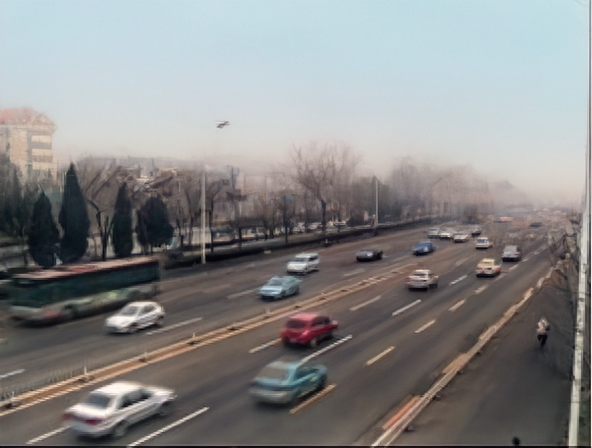}       
        \caption{ours (OTS)}
    \end{subfigure}
    \hfill
    \begin{subfigure}[b]{0.32\linewidth}
        \includegraphics[width=\linewidth]{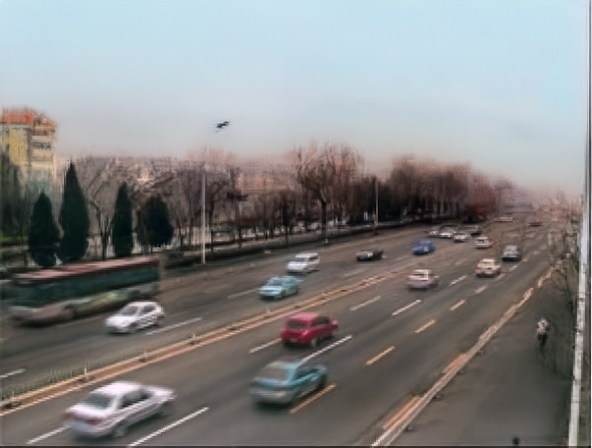}        
        \caption{ours (ITS)}
    \end{subfigure}
    \caption{Visual ablation results of the  IDDM.}
    \label{fig:abla}
\end{figure}
\begin{table}[t]
\centering
\small
\renewcommand{\arraystretch}{0.8}
\begin{tabular}{lcccc}
\toprule
Method & Brisque$\downarrow$ & NIQE$\downarrow$ & PIQE$\downarrow $& NIMA$\uparrow$ \\
\midrule
w/o IDDM & 23.726 & 4.783 & 36.263 & 4.465\\
w/o $\mathbf{h}_t$ & 24.777 & 3.879 & 35.973 & 4.492 \\
w/o Eq. \ref{eq:final_restoration} & 21.694 & 3.720 & 29.767 & 4.599  \\
ours & \textbf{18.321} & \textbf{3.494} & \textbf{27.072} & \textbf{4.840} \\
\bottomrule
\end{tabular}
\caption{Ablation study results trained on OTS using no-reference image quality metrics. ↑ indicates higher is better, ↓ indicates lower is better.}

\label{tab:ablation_noreference}
\end{table}

\begin{table}[t]
\centering
\small
\renewcommand{\arraystretch}{0.8}
\setlength{\tabcolsep}{2pt}
\begin{tabular}{l@{\hspace{0.8em}}ccc|l@{\hspace{0.8em}}ccc}
\toprule
\textbf{Setting} & {VI}$\uparrow$ & {RI}$\uparrow$ & {VSI}$\uparrow$ & {Setting} & {VI}$\uparrow$ & {RI}$\uparrow$ & {VSI}$\uparrow$ \\
\midrule
$\sigma \in [0.1,1.0]$ & 0.641 & 0.948 & 0.838 & {$T = 300$} &0.669 &0.905 &0.865 \\
$\sigma \in [0.4,1.5]$& {\textbf{0.879}} & {\textbf{0.959}}& {\textbf{0.936}}& {$T = 500$} & 0.750& 0.922 &0.881 \\
$\sigma \in [1.0,2.0] $& 0.616 & 0.911 & 0.822 &{$T = 1000$} & {\textbf{0.879}} & {\textbf{0.959} }& {\textbf{0.936}} \\
\bottomrule
\end{tabular}
\caption{Ablation studies on hyperparameters $\sigma$ (scattering coefficient range) and $T$ (diffusion timesteps) evaluated on the BEDDE dataset. Models are trained on OTS.}
\label{tab:ablation_beta_T}
\end{table}

\begin{table}[ht]
\centering
\small
\renewcommand{\arraystretch}{0.8}
\begin{tabular}{l@{\hspace{0.8em}}ccc}
\toprule
Setting & VI$\uparrow$ & RI$\uparrow$ & VSI$\uparrow$ \\
\midrule
$A \in [0.3,0.8]$ & 0.862 & 0.970 & 0.942 \\
$A\in [0.7,1.0] $& {\textbf{0.903}} & {\textbf{0.973}} & {\textbf{0.956}} \\
$A\in [0.9,1.2]$ & {0.882} & 0.971 & 0.943 \\
\bottomrule
\end{tabular}
\caption{Ablation studies on hyperparameter $A$ (atmospheric light range) evaluated on the BEDDE dataset. Models are trained on ITS.}
\label{tab:ablation_A}
\end{table}

It is evident that while other methods fail to clearly remove distant haze, IDDM effectively removes haze across the entire image, producing the best perceptual results in terms of colorfulness and brightness. Furthermore, due to both models trained on ITS and OTS can produce bright and clear dehazing images, the capability of domain generalization of our IDDM is proven.

\noindent \textbf{Extra Results on low-light enhancement}: Beyond our primary dehazing experiments, we find that when fed with low-light images without any specific retraining, IDDM produced visually enhanced results. Detailed visual results can be found in the Supplementary Material.

\subsection{Ablation Study}

\noindent \textbf{Studies of our framework components}:
Due to the intricate interdependencies within our training process, it is not feasible to present ablation results in a simple additive form (e.g., A, A+B, A+B+C). Therefore, our ablation study is organized into several comparative groups: 1) w/o IDDM, which only uses the normal diffusion model DDPM for image dehazing; 2) w/o $\mathbf{h}_t$, which means that you do not utilize $\mathbf{h}_t$ for noise predictor's training and sampling; 3) w/o Eq. \ref{eq:final_restoration}, which means directly applying simple normalization after sampling.
4) The results use indoor synthetic dataset (ITS) or outdoor synthetic dataset (OTS) for training. 

As shown in Fig. \ref{fig:abla} (b), without IDDM, the image remains hazy and unclear. In Fig. \ref{fig:abla} (c), the absence of $\mathbf{h}_t$ results in the color bias and distortion of the recovery image. In Fig. \ref{fig:abla} (e) and (f), the results trained with both indoor and outdoor datasets are provided, which shows that IDDM is robust for multi-domain generalization problems. 
We observe that in \ref{fig:abla} (e) and (f), the model trained on the ITS dataset removes more distant haze than that trained on the OTS dataset. This is because indoor environments often do not contain haze, whereas in natural scenes, a certain amount of haze is always present, and this cannot be avoided even in clear outdoor datasets. 
We also present qualitative results on RTTS \cite{reside} with all models trained on OTS, shown in Table \ref{tab:ablation_noreference}.

\noindent \textbf{Hyperparameter Study}:
 On the BEDDE dataset, we conducted comprehensive ablation studies on key hyperparameters while maintaining constant values for other variables ($T=1000$, $\sigma \in [0.4,1.5]$, and $A \in [0.7,1.0]$). As shown in Table \ref{tab:ablation_beta_T} and Table \ref{tab:ablation_A}, the experimental results demonstrate that the selected hyperparameters are optimal.

\section{Conclusion and Discussion}

In this paper, we present Image Dehazing Diffusion Models (IDDM), a novel approach addressing the challenge of synthetic-to-real domain discrepancy in image dehazing.  By integrating the atmospheric scattering model into diffusion frameworks, our method learns the physical process of haze formation during forward diffusion, enabling effective dehazing through sampling.

Extensive experiments demonstrate that IDDM consistently outperforms state-of-the-art methods across three real-world haze datasets. Notably, even when trained solely on ITS indoor synthetic dataset, our model achieves impressive dehazing results on real-world outdoor scenes. This cross-domain robustness stems from our incorporation of physical principles, which prevents artifacts and enables effective generalization across different visual domains.

{
    \small
    \bibliographystyle{ieeenat_fullname}
    \bibliography{main}
}


\end{document}